 \documentclass[tablecaption=bottom,wcp]{jmlr} 

\usepackage{booktabs}
\usepackage{wrapfig}
\usepackage[load-configurations=version-1]{siunitx} 

\theorembodyfont{\upshape}
\theoremheaderfont{\scshape}
\theorempostheader{:}
\theoremsep{\newline}

\newcommand{\argmax}{\operatornamewithlimits{argmax}}

\jmlrproceedings{AABI 2020}{3rd Symposium on Advances in Approximate Bayesian Inference, 2020}

\title[Annealed SVGD]{Annealed Stein Variational Gradient Descent}

 \author{\Name{Francesco D'Angelo} \Email{fdangelo@ethz.ch}\\
 \Name{Vincent Fortuin} \Email{fortuin@inf.ethz.ch}\\
 \addr ETH Zürich}

\begin{document}

\maketitle

\begin{abstract}
Particle-based optimization algorithms have recently been developed as sampling methods that iteratively update a set of particles to approximate a target distribution. In particular Stein variational gradient descent has gained attention in the approximate inference literature for its flexibility and accuracy. We empirically explore the ability of this method to sample from multi-modal distributions and focus on two important issues: (i) the inability of the particles to escape from local modes and (ii) the inefficacy in reproducing the density of the different regions. We propose an annealing schedule to solve these issues and show, through various experiments, how this simple solution leads to significant improvements in mode coverage, without invalidating any theoretical properties of the original algorithm. 
\end{abstract}


\section{Introduction}
There have been many recent advances on the theoretical properties of sampling algorithms for approximate inference, which changed our interpretation and understanding of them. Particularly worth mentioning is the work of \citet{jordan1998variational}, who reinterpret Markov Chain Monte Carlo (MCMC) as a gradient flow of the KL divergence over the Wasserstein space of probability measures. This new formulation not only allowed for a deeper understanding of these methods but also inspired the inception of new and more efficient inference strategies. Following this direction, \citet{liu2016stein} recently proposed the Stein Variational Gradient Descent (SVGD) to perform approximate Wasserstein gradient descent. This method belongs to the more general family of particle optimization variational inference (POVI), where a continuous density $p(x)$ is approximated by a set of $n$ particles that evolve over time towards the target. However, a solid understanding of its behavior in the finite particle limit beyond the mean field convergence analysis \citep{duncan2019geometry} remains elusive. What is more, there is empirical evidence that  SVGD suffers from a degeneracy that compromises the particle diversity under these conditions, making them collapse to a small number of modes \citep{zhuo2018message}. In the following, we discuss how an annealing strategy can significantly mitigate this issue, encourage exploration of the significant modes, and yield better samples from the target density than standard SVGD.

\subsection{Related work}
Introducing an artificial temperature parameter is a common practice in many approximate inference methods. Indeed, annealing approaches have been shown to be beneficial in both sampling and optimization problems for highly non-convex objectives. In the context of Markov chain Monte Carlo, sampling at different temperatures enhances the mixing times of the chains and thus allows for faster convergence \citep{marinari1992simulated, geyer1995annealing}. In Bayesian inference, a temperature parameter has been introduced to anneal the likelihood or the posterior of the model \citep[e.g.,][]{wenzel2020good} to escape from poor local minima. A similar effect can be obtained in variational inference and its stochastic counterpart by tempering the KL term \citep{mandt2016variational,huang2018improving, fu2019cyclical}. 
However, the impact of similar annealing strategies on the SVGD method have not yet been fully understood. Previous works \citep{chang2020kernel} studied the impact of a stochastic annealing introducing a sequence of noise-perturbed
score functions and \citet{han2018stein} proposed to perform SVGD on a series of intermediate tempered target distributions. In our work, we show that a deterministic annealing following a certain schedule can be useful to improve the particle diversity and overcome the mode-collapse problem.

\label{sec:intro}

\section{Background}
Stein variational gradient descent \citep{liu2016stein} is a technique to perform approximate inference using a set of particles $q_t(x) = \frac{1}{n} \sum_{i=1}^n \delta_{x_i(t)}$, with $\delta_{x_i}$ being the Dirac measure on particle $x_i$, to approximate a positive density function $p(x)$ on $\mathcal{X} \in \mathbb{R}^d$. More precisely, SVGD is an efficient numerical technique to discretize the Wasserstein gradient flow of the Kullback-Leibler (KL) divergence functional on a new metric called the Stein geometry \citep{duncan2019geometry}.

SVGD considers an incremental transformation given by an infinitesimal perturbation of the identity matrix $\mathbf{T}(x) = x + \epsilon \phi(x)$ to move the particles from the initialization to the target. Here, $\phi(x)$ is the direction of the perturbation and $\epsilon$ the step size. The former is chosen to maximally decrease the KL divergence between the discrete density of the particles and the final target. As shown in \citet{liu2016stein}, closed-form solutions can be obtained when restricting all perturbations $\phi$ to be from the unit ball of a vector valued reproducing kernel Hilbert space (RKHS) $\mathcal{H}^d = \mathcal{H}_0 \times ... \times \mathcal{H}_0$.
Here, $\mathcal{H}_0$ is a scalar-valued RKHS associated with a scalar positive definite kernel $k(x,x';h)$, and $h$ is the set of kernel hyperparameters. The direction of steepest descent that maximizes the negative gradient of the KL divergence is then given in closed form as: 
\begin{equation}
    \phi_{q,p}^*(x') = \argmax_\phi \bigg\{ -\nabla_\epsilon D_{KL}(q_{[\mathbf{T}]}||p) \big|_{\epsilon \to 0} \bigg\} \propto \mathbb{E}_{x \sim q} [\mathcal{A}_p k(x,x')] \, ,
    \label{eqn:opt_pert}
\end{equation}
with $\mathcal{A}_p \phi(x) = \phi(x) \nabla_x \log p(x)^\top + \nabla_x \phi(x)$ being the Stein operator.
Using this, we can build an iterative procedure that transforms the initial reference distribution $q_0$ to the target posterior. Practically, we draw a set of particles $\{ x^0_i\}_{i=1}^n$ with $x^0_i \sim q_0$ and subsequently update them using the optimal perturbation in \eqref{eqn:opt_pert}:
\begin{equation}
    x_i^{t+1} \leftarrow x_i^t + \epsilon_t \hat{\phi}^* (x_i^t) \ \ \ \text{with} \ \ \ \hat{\phi}^*(x) = \frac{1}{n} \sum_{j=1}^n [ \underbrace{k(x_j^t,x) \nabla_{x_j^t} \log p(x_j^t)}_{\text{driving force}} + \underbrace{\nabla_{x_j^t} k(x_j^T,x)}_{\text{repulsive force}}]
    \label{eqn:update}
\end{equation}
Conceptually the update rule attracts the particles to high density regions of the target via the average score function (driving force in \eqref{eqn:update}) while the repulsive force pushes them away from each other. This avoids a collapse to the MAP estimate and allows a certain degree of diversity among the particles to encourage the exploration of multiple modes and a more faithful reflection of the variance of the target distribution.

\subsection{Mode-collapse in SVGD}
\label{sec:pitfall}

\begin{wrapfigure}{l}{0.40\textwidth}
\begin{center}
\vspace{-5mm}
\includegraphics[width=0.40\textwidth]{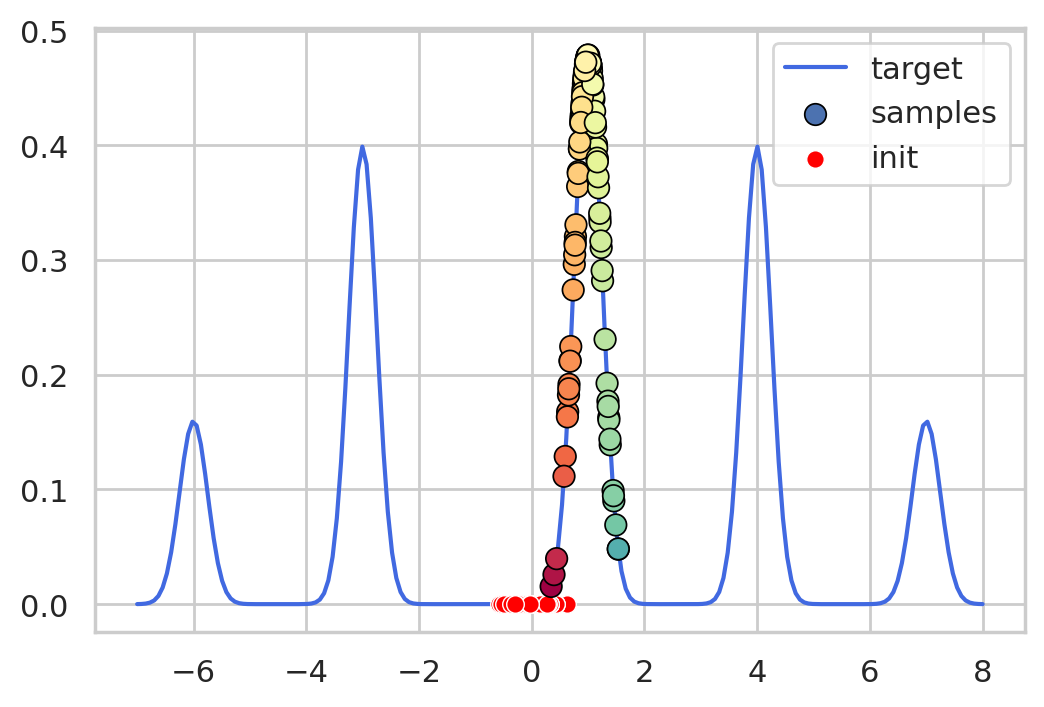}
\includegraphics[width=0.40\textwidth]{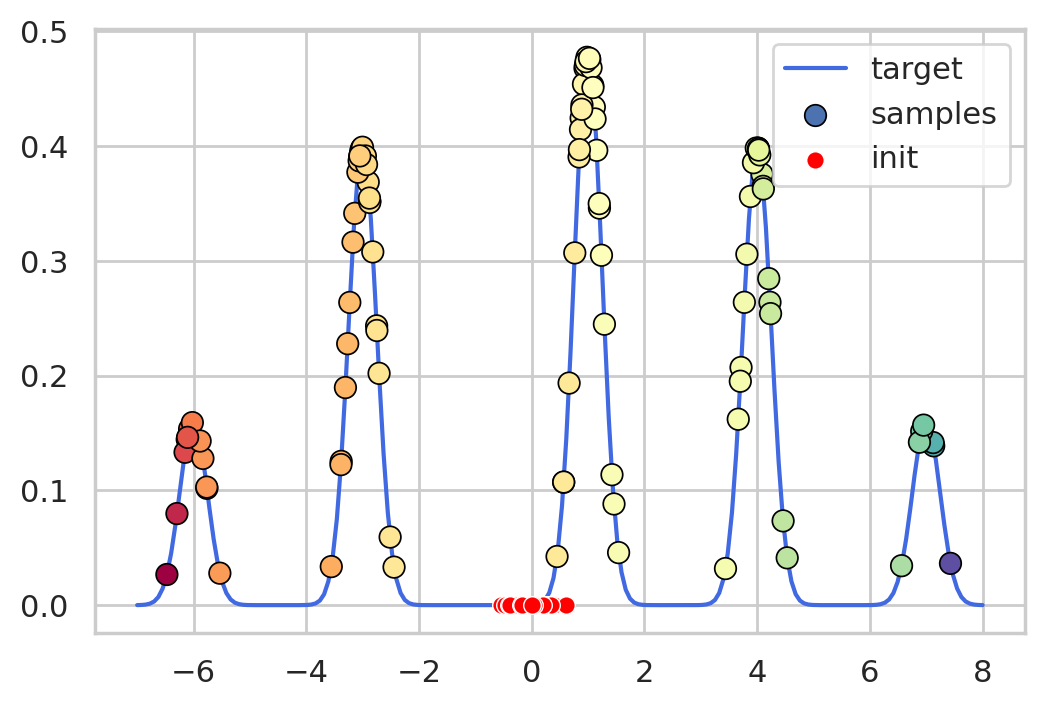}
\end{center}
\caption{\textbf{SVGD mode-collapse.} Comparison of SVGD (top) and our proposed A-SVGD (bottom).}
\label{fig:1d_gaus_collapse}
\end{wrapfigure}

\begin{figure*}
\centering
\includegraphics[width=1.0\textwidth]{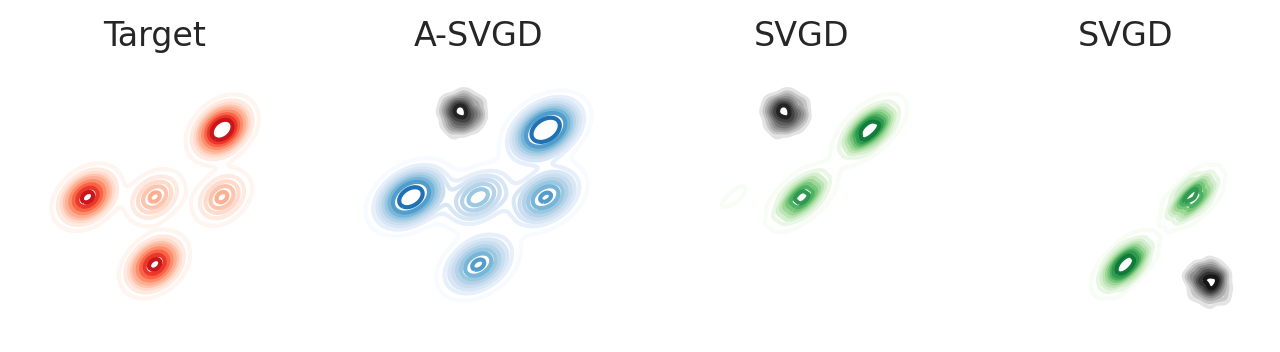}
\caption{\textbf{Mode covering of SVGD.} We compare the final stationary distribution of standard SVGD (green) from two different initialization (black) and A-SVGD (blue) to approximate a mixture of Gaussians (red).}
\label{fig:modes_covering}
\end{figure*}

Despite its theoretical foundations, it has been shown empirically \citep{zhuo2018message} and theoretically \citep{zhang2020stochastic} that the particles in SVGD tend to collapse to a few local modes and that this effect is strongly related with the initial distribution of the particles. This issue is also clearly visible in our experiments and seems to already be relevant in low-dimensional problems. Indeed, as shown in Figure~\ref{fig:1d_gaus_collapse}, it even happens in the case of a one-dimensional mixture of five Gaussians. Here, all particles, independent of the choice of the kernel bandwidth (see Appendix~\ref{apd:second}),  end up in the mode closest to the initialization without any possibility of escaping.
\newline
Additionally, we noticed in our experiments a connection of how a proper reconstruction of the target mass in the different modes is related to the initialization, as seen in Figure~\ref{fig:modes_covering}. In other words, even if the standard SVGD is able to capture different modes, the particles are not correctly distributed to faithfully reproduce the original mass. Instead, the majority of them tend to end up in the mode closest to their initialization. It is not clear whether this particular issue is a consequence of the mode-seeking limitation of all KL-divergence-based inference methods or of the approximation due to a finite number of particles. Empirically evident instead is that a deterministic update of the samples, like the one characterizing SVGD, in combination with a random initialization, can lead to a catastrophic convergence. In comparison, other methods characterized by a stochastic update like SGLD \citep{welling2011bayesian} can instead always rely on a nonzero probability of escaping from a certain mode, given by the injected noise, and consequently mitigate a bad initialization and encourage exploration. Moreover, it has been shown by \citet{zhang2020stochastic} that injecting noise is also beneficial for the SVGD update to overcome the mode-collapse issues.
Finally, it is well established that distance metrics and corresponding kernel-based methods suffer from the curse of dimensionality \citep{aggarwal2001surprising,reddi2014decreasing}. That is why we should not hope for this problem to disappear in higher dimensions, but rather expect it to become even worse.

\section{Annealed SVGD}
Motivated by related work that mitigates similar mode-collapse issues in MCMC methods by tempering the target density to speed up the mixing time and avoid chains to get stuck in a single mode \citep{neal1996sampling}, we propose to introduce an annealing schedule in the SVGD update (A-SVGD). This modification keeps the deterministic nature of the method but is essential to enhance its exploration and mode coverage and ideally would compensate for the limitation of the initialization and the finite number of particles. We introduce an annealing parameter $\gamma(t) \in [0,1]$ depending on the current iteration step and modify the update rule from \eqref{eqn:update} in the following way: 
\begin{equation}
    \hat{\phi}^*(x) = \frac{1}{n} \sum_{j=1}^n [ \underbrace{ \gamma(t)k(x_j^t,x) \nabla_{x_j^t} \log p(x_j^t)}_{\text{driving force}} + \underbrace{\nabla_{x_j^t} k(x_j^T,x)}_{\text{repulsive force}}] \, .
\end{equation}
Intuitively, we can observe two phases in the time evolution of the particles by varying $\gamma$ in the interval $[0,1]$ with an appropriate schedule: The first phase is exploratory with a predominant repulsive force that pushes the particles away from the initialization and thus allows for a good coverage of the target distribution's support. The second phase is exploitative, where the driving force takes over and shrinks the distribution of the particles to the area around the different modes. From a statistical perspective, our modification corresponds to the introduction of a temperature parameter $T(t) = \frac{1}{\gamma(t)}$ which rescales the target distribution $p(x)^\frac{1}{T(t)}$ during the evolution of the approximating density. It is important to notice how the choice of the annealing schedule is fundamental to preserve the convergence properties of SVGD and to keep the final target density unchanged. We ensure this by formulating the annealing schedule in such a way that the final iterations are always performed for the true target density, that is, $\lim_{t \to \infty} \gamma(t) = 1$. From this point of view, our alternative method is formally equivalent to a better parametrization of the initial reference distribution $q_0$ of the particles that depends on the target distribution. This is due to the fact that even if in the exploratory phase, when the repulsion is dominant, we still have a small component of the driving force that ensures that the particles are not randomly driven far away from the initialization, but are still driven towards high-density regions.
%
%
%
\subsection{The annealing schedule}
\label{sec:ann_schedule}
\begin{figure*}
\centering
\includegraphics[width=0.45\textwidth]{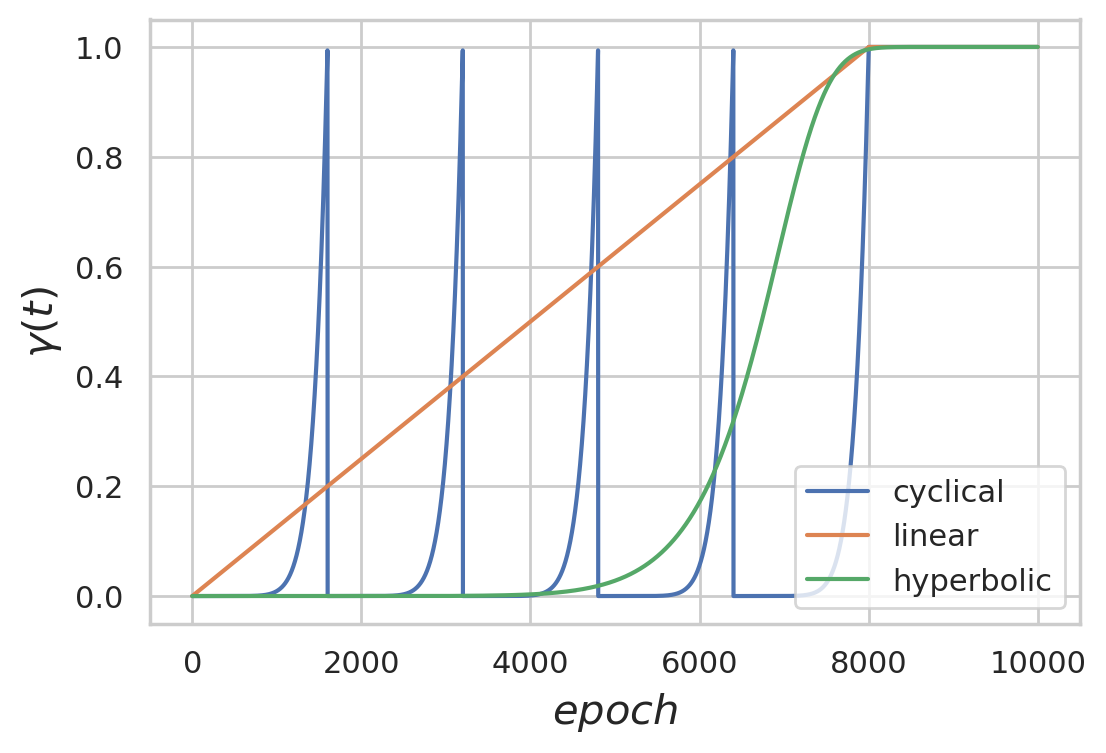}
\includegraphics[width=0.45\textwidth]{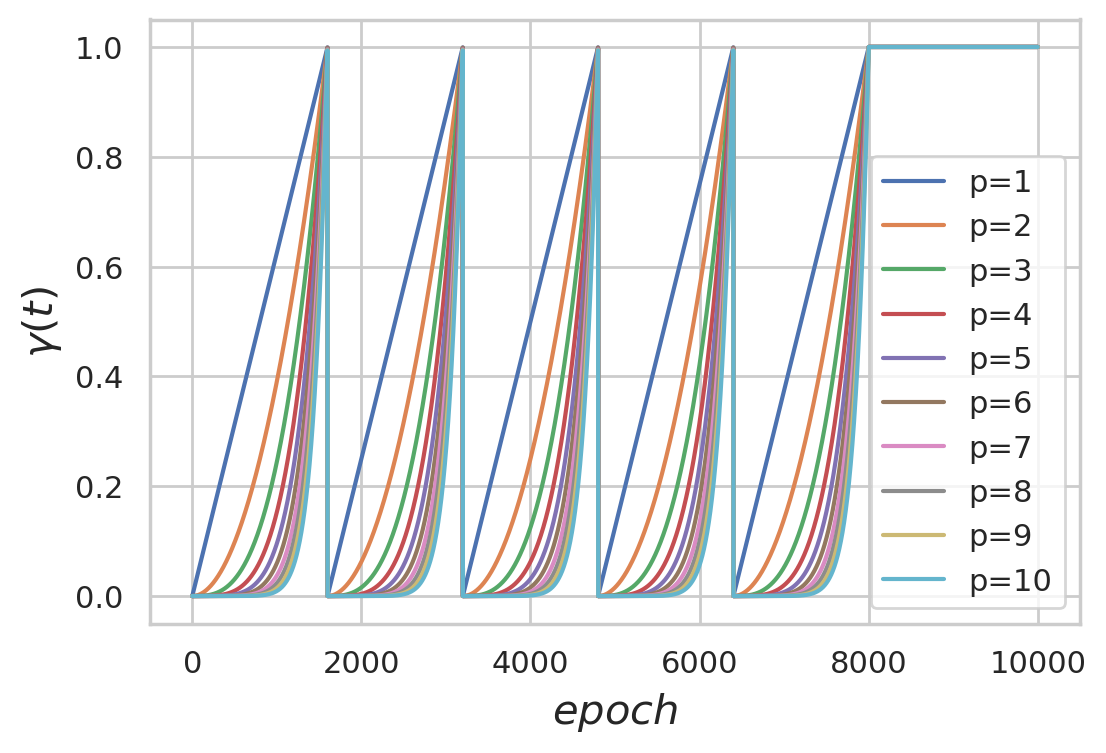}
\caption{\textbf{Annealing schedules.} An illustration of the proposed annealing schedules.}
\label{fig:ann_schedule}
\end{figure*}

As mentioned before, the choice of the annealing schedule is fundamental for the convergence to multiple modes while yielding samples from the proper target distribution. For this purpose, we introduced and tested different annealing schedules as shown in Figure~\ref{fig:ann_schedule}. The simplest idea is a linear annealing on the interval $[0,1]$; however, our experiments showed that this choice is not optimal. Indeed, linear tempering of the density leads to slow particle dynamics that are beneficial to neither the exploration nor the convergence. For this reason we chose to make the transition between the two inference phases steeper, so that the majority of the evolution happens in one of the two phases and not in between. To do so, we construct the annealing schedule using the hyperbolic tangent: $\gamma(t) = \tanh \big[(1.3 \frac{t}{T})^{p}\big]$, with $t$ being the current time step and $T$ the total number of steps. Despite its good exploration, this method might encourage modes very far from the initialization due to the flattening of the target density in the initial exploratory phase. To achieve a tradeoff and have a more ``target-guided'' exploration, we follow the cyclical annealing schedule idea proposed in \citet{loshchilov2016sgdr} and \citet{huang2017snapshot}, which has already been used for MCMC sampling in \citet{zhang2019cyclical}. We adapted this technique to have a sequence of $C$ cycles of exploratory and converging phases, obtaining the following expression: 

\begin{equation}
        \gamma(t) = \bigg( \frac{mod(t, T/C)}{T/C} \bigg)^p \, ,
    \label{eqn:cyclic_ann}
\end{equation}
with $T$ being the total number of time steps, $C$ the number of cycles and $p$ an exponent determining the  speed of the transition between the two phases (as shown in Figure~\ref{fig:ann_schedule}).

\section{Experiments}
\label{sec:exp}
We demonstrate the advantages of our method in several synthetic experiments. In all the experiments we used SVGD with a standard RBF kernel.

\paragraph{Univariate Gaussian mixture.}

We first assessed the ability of A-SVGD to sample from a multi-modal univariate distribution given by a mixture of five Gaussians.
The step size was fixed to $\epsilon = 0.1$ and we used the hyperbolic annealing schedule.
We see in Figure~\ref{fig:1d_gaus_collapse} that our proposed method successfully covers all the modes, while the standard SVGD collapses to just a single mode.

\begin{figure*}
\centering
\includegraphics[width=0.40\textwidth]{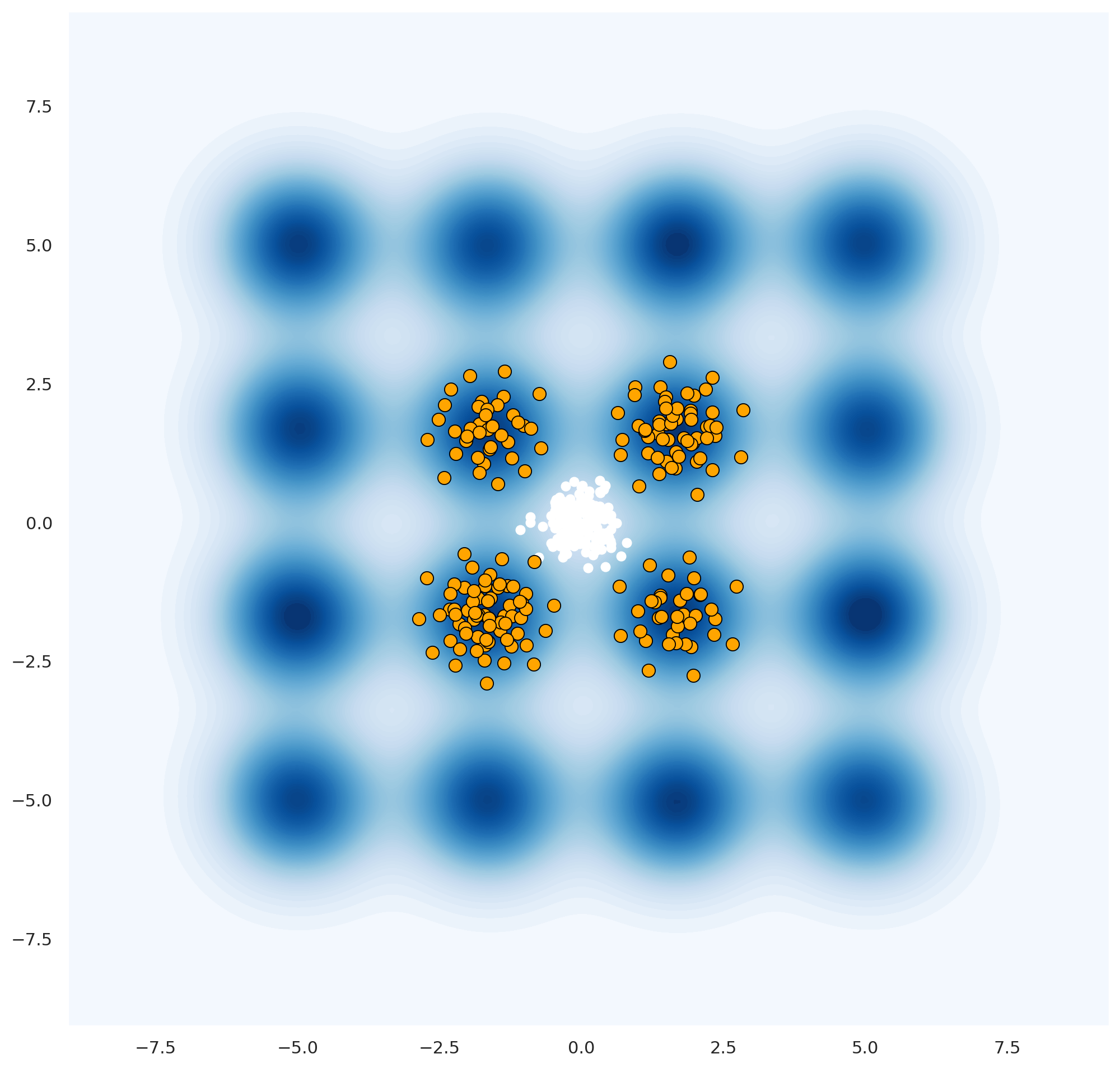}
\includegraphics[width=0.40\textwidth]{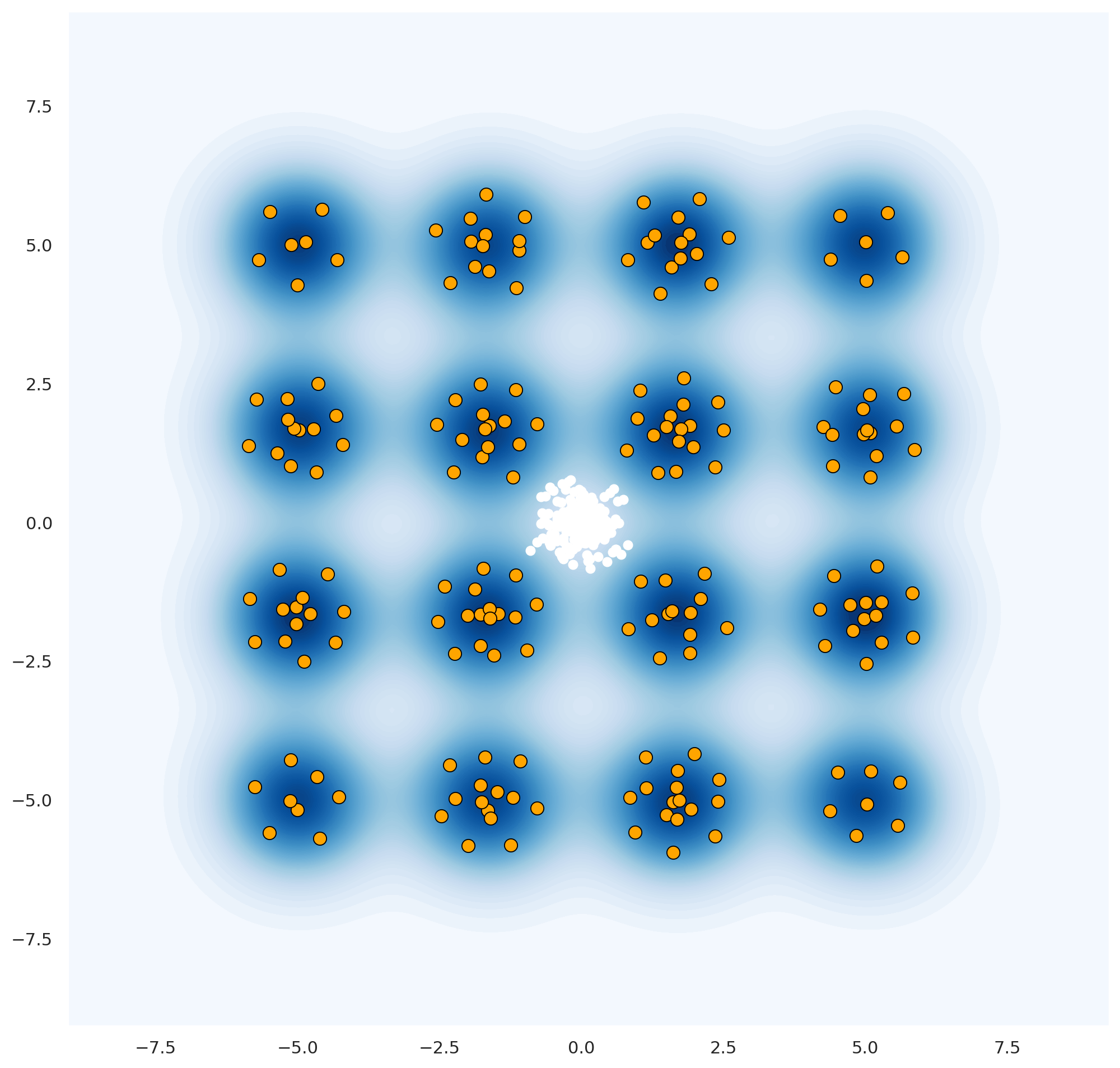}
\caption{\textbf{SVGD on multi-modal 2D data.} We show the final samples of SVGD with annealing (right) and without (left) starting from the same initial distribution (white dots).  }
\label{fig:multi_data}
\end{figure*}

\paragraph{Bivariate regular Gaussian mixture.}
Secondly, we tested our method on a 2D mixture of 16 Gaussians with means equally distributed on a $4 \times 4$ grid and standard deviation $\sigma = 0.5$.
In this experiment we used the cyclical annealing schedule from \eqref{eqn:cyclic_ann}.
As reported in Figure~\ref{fig:multi_data}, we observe that the standard SVGD gets trapped in four of the modes, neighboring the initialization. In contrast, our method is able to find and characterize all modes, independently of the initial position.

\paragraph{Bivariate irregular Gaussian mixture.}
In our last experiment we studied the ability of SVGD to reproduce the weights of the mixture components of 2D Gaussians.
The cyclical annealing schedule is used for the A-SVGD and for the standard SVGD we show two different initializations to illustrate their impact.
We see in Figure~\ref{fig:multi_data} that our proposed method not only covers all the modes, but also approximately recovers their mixture weights, while the standard SVGD again collapses to the modes that are closest to its initialization.

\paragraph{Additional results.}
We provide additional results and ablation studies in the appendix.
We show in appendix~\ref{apd:first} that even when the particles are initialized exactly in one mode of the Gaussian mixture, the A-SVGD can cover all modes, while the standard SVGD stays trapped in that single mode.
Moreover, we show in appendix~\ref{apd:second} that changing the bandwidth of the RBF kernel cannot overcome the mode collapse, in contrast to our proposed annealing.
Appendix~\ref{apd:third} shows that the cyclical annealing scheme generally outperforms the other ones and appendix~\ref{apd:fourth} confirms our findings also in high-dimensional settings.

\section{Conclusion}

In this work, we discussed the mode-collapse issue of SVGD for approximate inference and proposed an annealing strategy to overcome these limitations. We illustrated the impact of the initialization on a deterministic sampling algorithm like SVGD, highlighting two major drawbacks, namely (i) a tendency of the particles to fall into the neighboring modes without any possibility of escape and (ii) the difficulty of the particles in reproducing the effective local density of any given mode. We found that the introduction of a temperature parameter and an annealing schedule can help alleviate these undesirable behaviors, leading to better samples that can effectively capture multi-modal densities.  





\bibliography{jmlr-sample}
\newpage 
\appendix

\counterwithin{figure}{section}
\counterwithin{table}{section}

\section{Additional results on multi-modal data}\label{apd:first}
\begin{figure*}[h]
\includegraphics[width=0.50\textwidth]{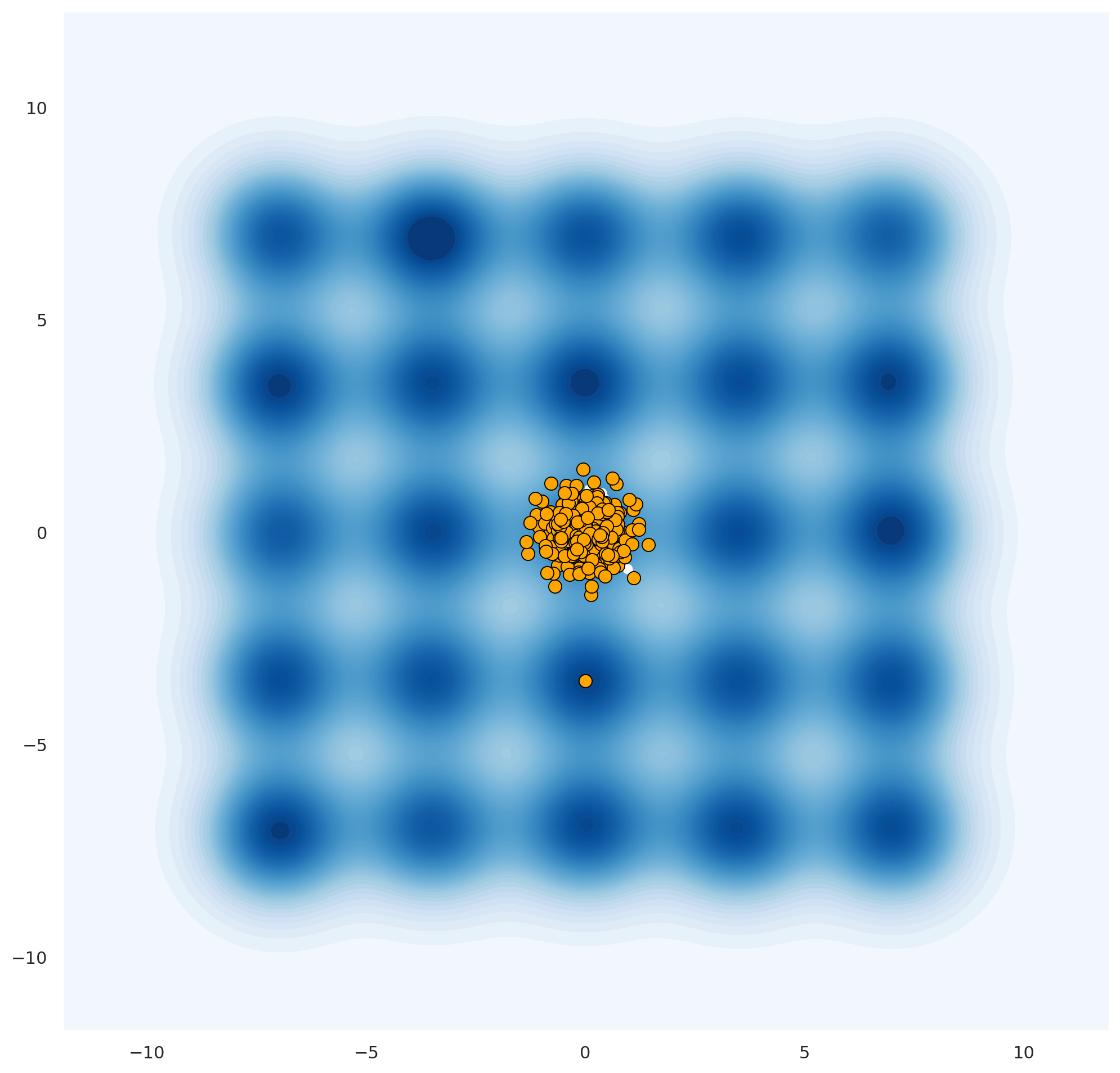}
\includegraphics[width=0.50\textwidth]{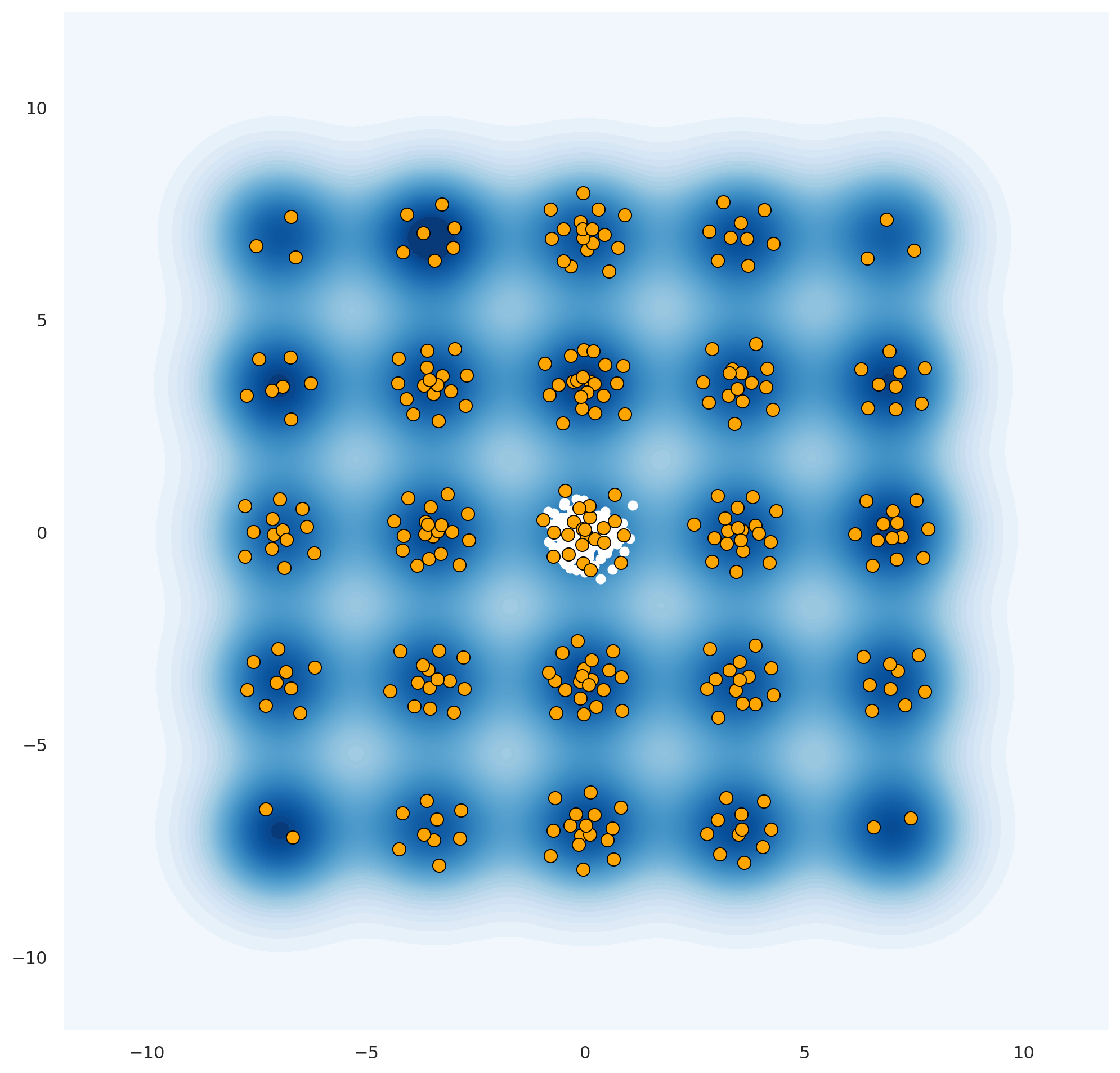}
\caption{\textbf{SVGD on multi-modal data with central initialization.} We show the final samples of SVGD with annealing (right) and without (left) starting from the same initial distribution (white dots). In this particular case the particles are initialized in the central mode. This particular initialization perfectly shows how the standard SVGD is not able to efficiently cover the entire target density, but instead remains trapped in the initialization mode.}
\label{fig:multi_data_cent}
\end{figure*}

In extension to the multi-modal synthetic data presented in section \ref{sec:exp} we present here the extreme case for which the initialization is exactly in one of the modes of the target distribution. As shown in Figure~\ref{fig:multi_data_cent}, the SVGD with annealing is remarkably able to escape from the initialization, covering all the modes characterizing the target density. On the other hand, the standard SVGD is not able to model anything besides the mode in which the particles have been initialized. 
\section{Different RBF kernel bandwidths }\label{apd:second}
\begin{figure*}
\centering
\includegraphics[width=1.0\textwidth]{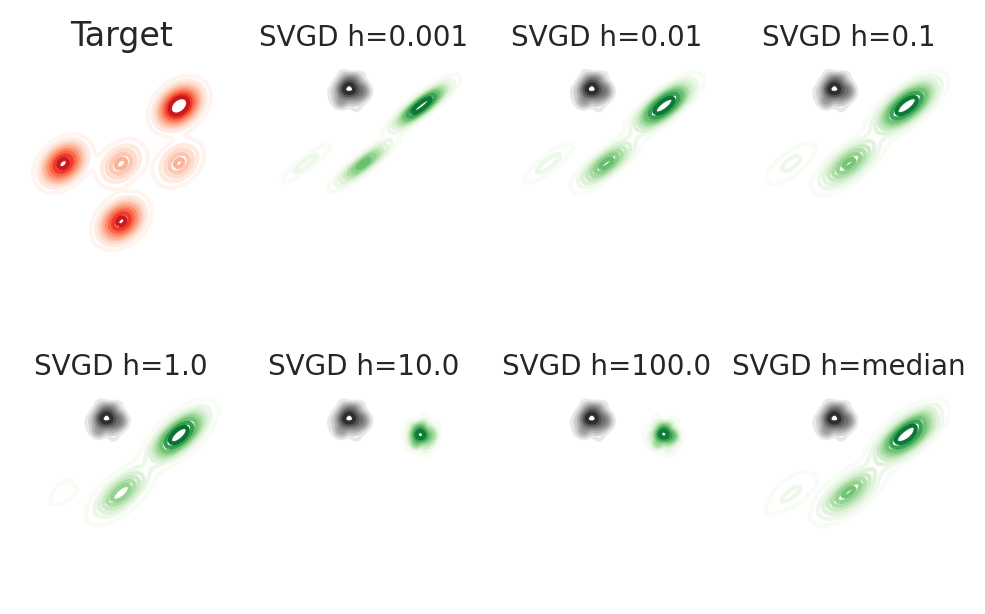}
\caption{\textbf{Mode covering of SVGD for different bandwidth.} We compare the final stationary distribution of standard SVGD (green) using different bandwidth (h). }
\label{fig:bandwidth_effect}
\end{figure*}

We compared the effect of different bandwidths for the RBF kernel to assess if this parameter affects the issues illustrated in \ref{sec:pitfall}. We used the synthetic multi-modal data from Figure~\ref{fig:modes_covering} to test the following bandwidth values $h \in \{0.001, 0.01, 0.1, 1, 10,100, \text{median}\}$ where median is the median heuristic. As illustrated in Figure~\ref{fig:bandwidth_effect}, none of the bandwidths shows significant improvement and, in contrast to our A-SVGD, the inference is still limited to the neighboring modes. 

\section{Different annealing schedules }\label{apd:third}
\begin{figure*}[h]
\centering
\includegraphics[width=1\textwidth]{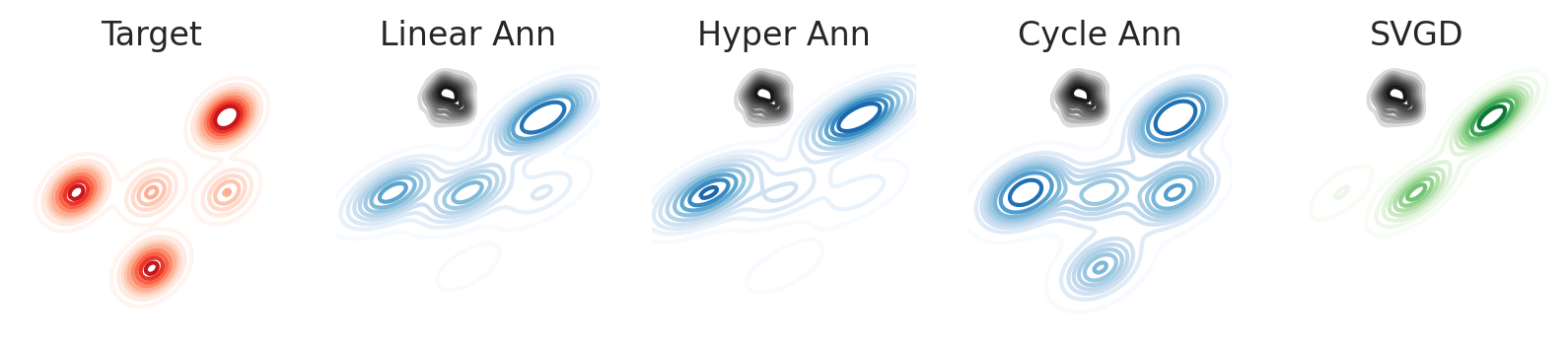}
\caption{\textbf{Mode  covering  of different annealing schedules.}We  compare  the  final  stationary  distribution  of standard  SVGD  (green) and  A-SVGD (blue) using the three different annealing schedules to approximate a mixture of Gaussians (red) }
\label{fig:modes_cov_ann}
\end{figure*}

\begin{figure*}[h]
\centering
\includegraphics[width=0.7\textwidth]{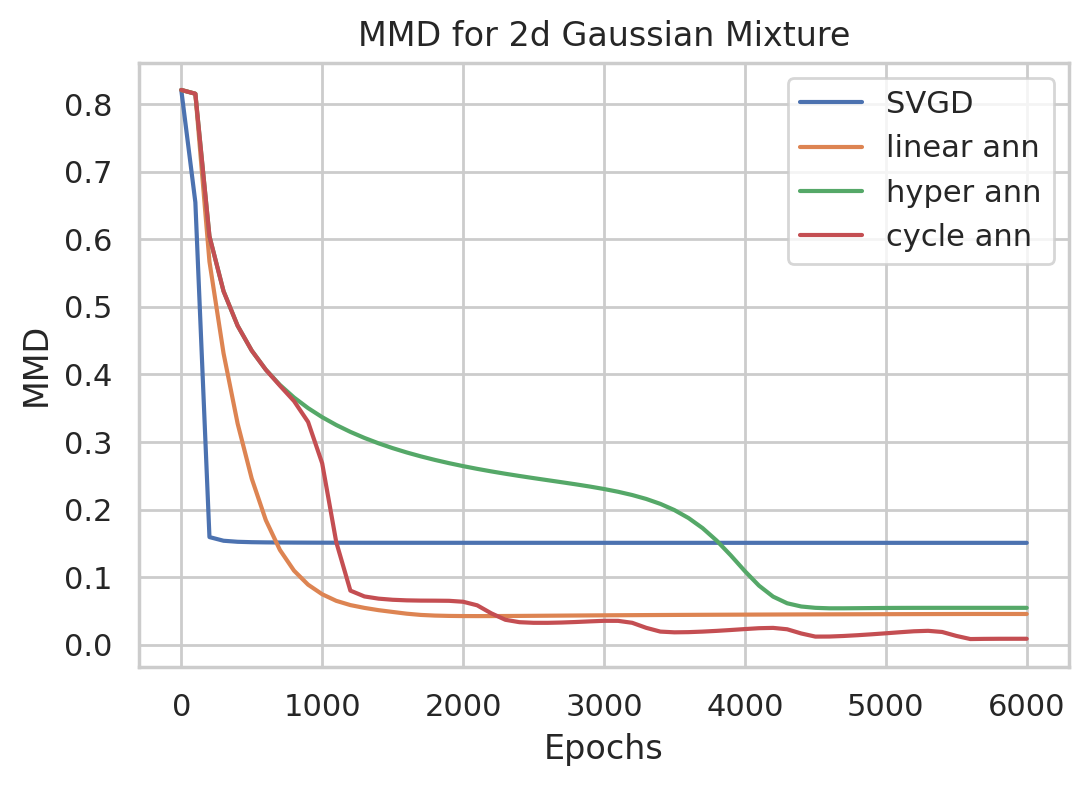}
\caption{\textbf{MMD for different annealing schedules.} We compute MMD during the evolution of the particles for the three different proposed annealing schedules }
\label{fig:MMD_ann}
\end{figure*}

We compared the different annealing schedules proposed in section \ref{sec:ann_schedule} on the bivariate irregular Gaussian mixture as reported in Figure~\ref{fig:modes_cov_ann}. We also computed the maximum mean discrepancy (MMD) \citep{gretton2012kernel} during the evolution of the particles as reported in Figure~\ref{fig:MMD_ann}.

\section{High-dimensional experiment}\label{apd:fourth}
To assess the capability of the annealed SVGD in high-dimensional settings, we studied its ability to produce samples from a mixture of 5 Gaussians in $d = 100$ dimensions. We sampled the means of the 5 components of the mixture according to $\mu_i \sim \mathcal{N}(0,4\mathbb{I})$ so that they would be distributed on the annular region at distance $\propto \sqrt{d}$ from the origin. We initialize $5000$ particles sampling from $x_0 \sim \mathcal{N}(0,\mathbb{I})$ and evolved them for $120k$ iterations with a stepsize $\epsilon = 0.3$. We evaluate convergence of the particles to the target distribution measuring the MMD every 100 iterations for the different schedules.
\begin{figure*}[h]
\centering
\includegraphics[width=0.7\textwidth]{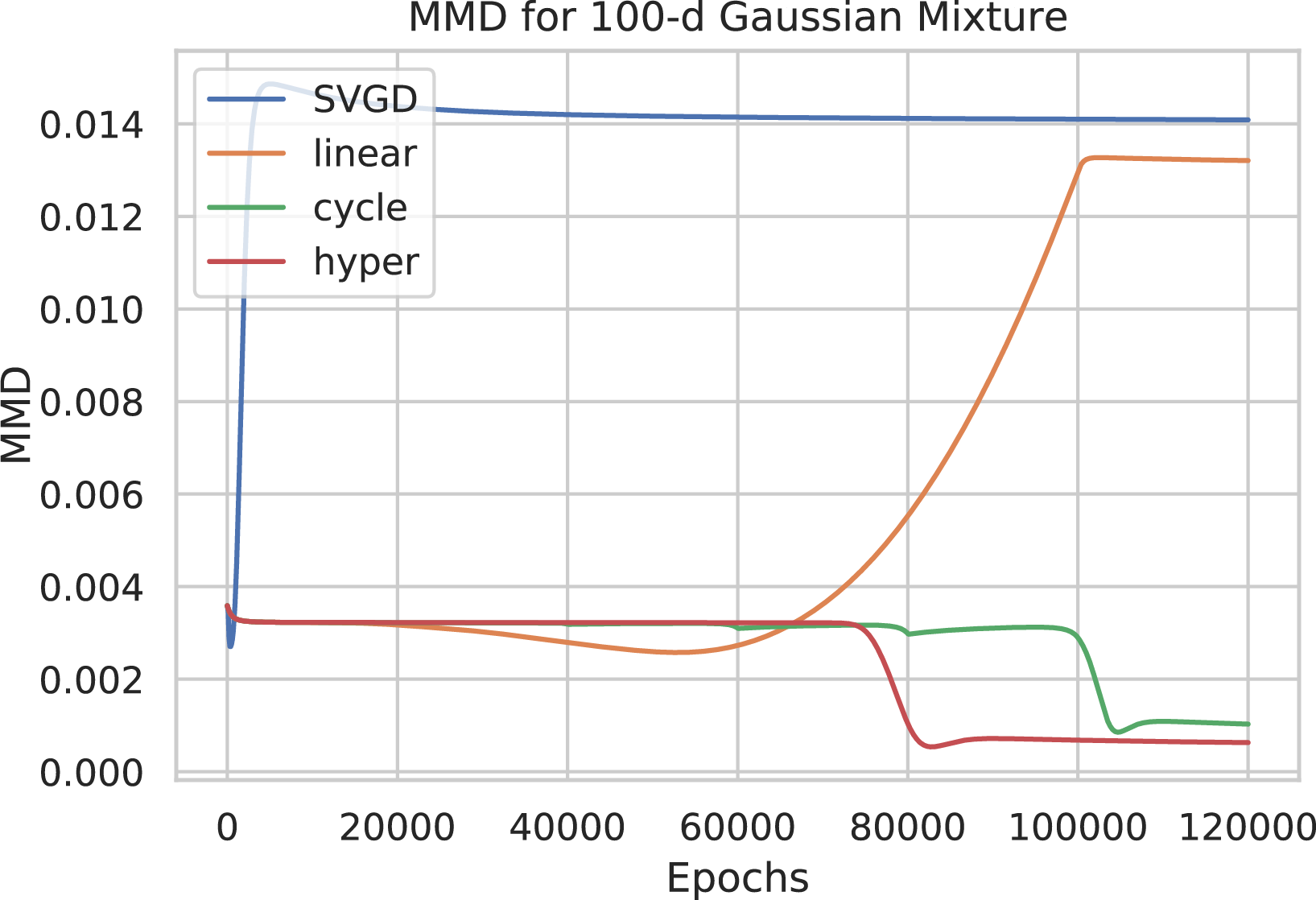}
\caption{\textbf{MMD for 100 dimensional Gaussian mixture.} We compute MMD during the evolution of the particles for the three different proposed annealing schedules when the target distribution is a mixture of five Gaussians in 100 dimensions.}
\label{fig:MMD_ann_100}
\end{figure*}
Additionally we studied the coverage of the different modes by measuring the distance of the samples from them. Thereby we can determine whether or not the particles are only collapsing into the closest mode to the initialization. Moreover, we can establish if they are able to faithfully reproduce the distance of the samples from the modes.

\begin{figure*}
\centering
\includegraphics[width=1.0\textwidth]{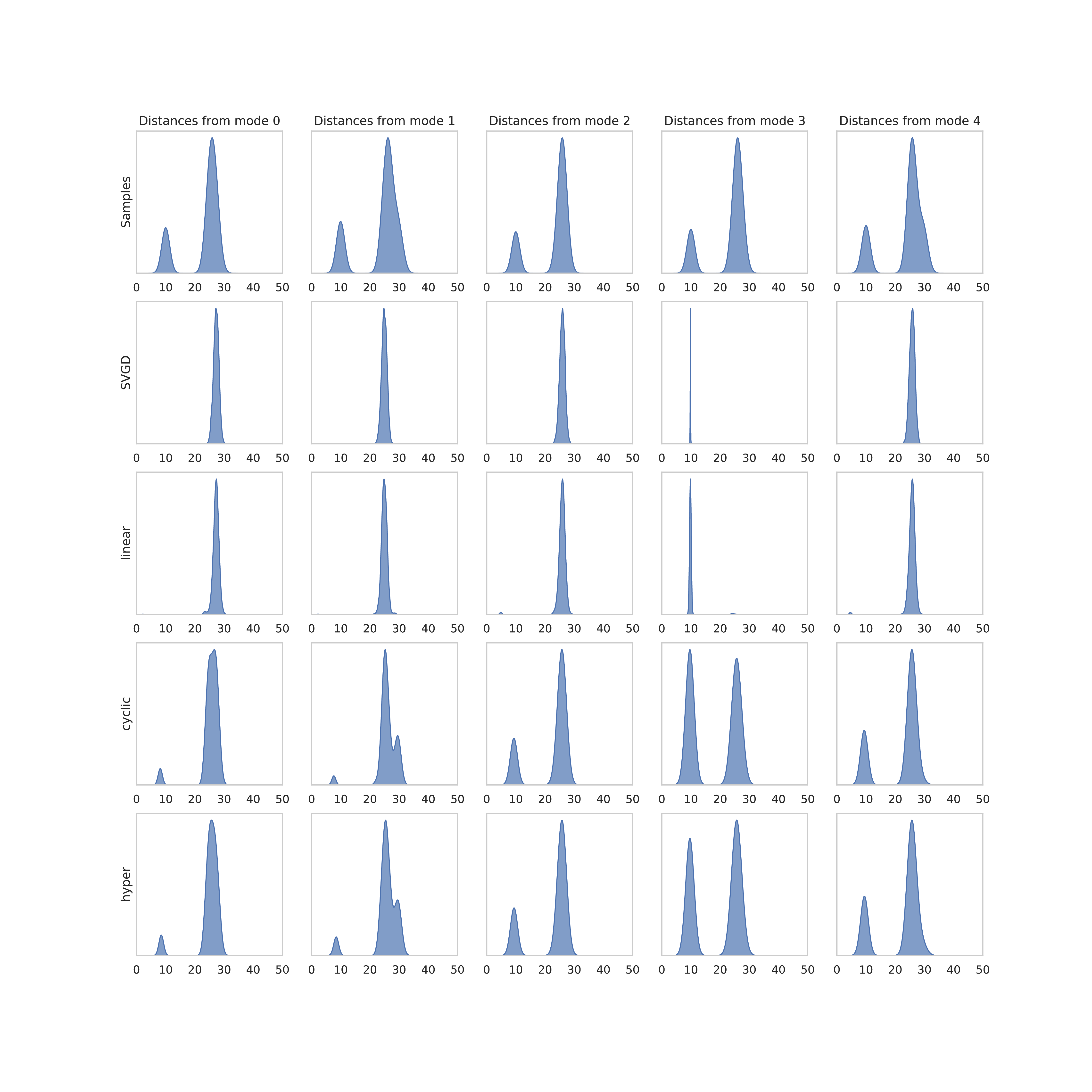}
\caption{\textbf{Distances from the modes.} We compare the distance of the samples to the modes for SVGD with and without annealing when the target is a 100 dimensional Gaussian mixture. }
\label{fig:distances_100d_mixture}
\end{figure*}

The results are reported in Figure~\ref{fig:MMD_ann_100}. Interestingly, we can see how the standard SVGD completely fails in modelling this high-dimensional distribution indeed the MMD, due to the collapse to a single mode, increases during the iterations converging to a unimodal final distribution. Conversely the introduction of an annealing schedule, in particular the hyperbolic and cyclical ones, allows for a more efficient exploration that avoids collapsing to a single mode and thus decreases the MMD. Additionally, the histograms reported in Figure~\ref{fig:distances_100d_mixture} show the distribution of the distances of the particles from the single modes and highlight that for standard SVGD and linear annealing, the particles indeed collapse to a single mode, while for hyperbolic and cyclical annealing, they cover all the modes.  

\end{document}